# Using New Data to Refine a Bayesian Network*

**Wai Lam and Fahiem Bacchus**
Department of Computer Science
University of Waterloo
Waterloo, Ontario,
Canada, N2L 3G1

## Abstract

We explore the issue of refining an existent Bayesian network structure using new data which might mention only a subset of the variables. Most previous works have only considered the refinement of the network's conditional probability parameters, and have not addressed the issue of refining the network's structure. We develop a new approach for refining the network's structure. Our approach is based on the Minimal Description Length (MDL) principle, and it employs an adapted version of a Bayesian network learning algorithm developed in our previous work. One of the adaptations required is to modify the previous algorithm to account for the structure of the existent network. The learning algorithm generates a partial network structure which can then be used to improve the existent network. We also present experimental evidence demonstrating the effectiveness of our approach.

## 1 Introduction

A number of errors and inaccuracies can occur during the construction of a Bayesian net. For example, if knowledge is acquired from domain experts, miscommunication between the expert and the network builder might result in errors in the network model. Similarly, if the network is being constructed from raw data, the data set might be inadequate or inaccurate. Nevertheless, with sufficient engineering effort an adequate network model can often be constructed. Such a network can be usefully employed for reasoning about its domain. However, in the period that the network is being used, new data about the domain can be accumulated, perhaps from the cases the network is used on, or from other sources. Hence, it is advantageous to be able to improve or refine the network using this new data. Besides improving the network's accuracy, refinement can also provide an adaptive ability. In particular, if the probabilistic process underlying the relationship between the domain variables changes over time, the network can be adapted to these changes and its accuracy maintained through a process of refinement. One way the underlying probabilistic process can change is via changes to variables not represented in the model. For example, a model of infectious disease might not include variables for public sanitation, but changes in these variables could well affect the variables in the model.

A number of recent works have addressed the issue of refining a network, e.g., [Bun91, Die93, Mus93, SL90, SC92]. However, most of this work has concentrated on improving the conditional probability parameters associated with the network. There has been very little work on actually improving the *structure* of the network. Clearly errors in the construction of the original network could have just as easily yielded inaccurate structures as inaccurate probability parameters. Similarly, changes to variables not included in the network could result in structural as well as parametric changes to the probabilistic process underlying the network variables. This paper presents a new approach to the problem of structural refinement. The approach uses new data, which might be only partially specified, to improve an existent network making it more accurate and more useful in domains where the underlying probabilistic process changes over time.

It is not uncommon that practical Bayesian network structures involve a fairly large number (e.g., $> 100$) of nodes. However, inaccuracies or changes might affect only some subset of the variables in the network. Furthermore, new data collected about the domain might be only partial. That is, the data might contain information about only a subset of the network variables. When the new data is partial it is only possible to refine the structural relationships that exist between the variables mentioned in the new data. Our approach performs refinement locally; i.e., it uses an algorithm that only examines a node and its parents. This allows us to employ partial data to improve local sections of the network. Furthermore, by taking advantage of locality our approach can avoid the potentially very expensive task of examining the entire network.

*Wai Lam's work was supported by an OGS scholarship. Fahiem Bacchus's work was supported by the Canadian Government through the NSERC and IRIS programs. The authors' e-mail addresses are [wlam1|fbacchus]@logos.uwaterloo.ca.

Our approach makes use of the Minimal Description Length (MDL) principle [Ris89]. The MDL principle is a machine learning paradigm that has attracted the attention of many researchers, and has been successfully applied to various learning problems, see, e.g., [GL89, QR89, LB92]. Specifically, we adapt the MDL learning algorithm developed in our previous work [LB93] to the refinement task, and perform experiments to demonstrate its viability.

In the subsequent sections we first present, more formally, the problem we are trying to solve. After an overview of our approach we show how the MDL principle can be applied to yield a refinement algorithm. This requires a further discussion of how the component description lengths can be computed. The refinement algorithm is then presented. Finally, we present some results from experiments we used to evaluate our approach. Before, turning to these details, however, let us briefly discuss some of the relevant previous work on refinement.

Spiegelhalter et al. [SL90, SC92] developed a method to update or refine the probability parameters of a Bayesian networks using new data. Their method was subsequently extended by other researchers [Die93, Mus93]. However, all of these approaches were restricted to refining the probability parameters in a fixed network. In other words, they were not capable of refining or improving the network's structure. Buntine [Bun91] proposed a Bayesian approach for refining both the parameters and the structure. The initial structure is acquired from prior probabilities associated with each possible arc in the domain. Based on the new data, the structure is updated by calculating the posterior probabilities of these arcs. Buntine's approach can be viewed as placing a prior distribution on the space of candidate networks which is updated by the new data. One can also view the MDL measure as placing a prior distribution on the space of candidate networks. However, by using an MDL approach we are able to provide an intuitive mechanism for specifying the prior distribution that can be tailored to favour preserving the existent network to various degrees. Finally, Cowell et al. [CDS93] have recently investigated the task of *monitoring* a network in the presence of new data. A major drawback of their approach that it can only detect discrepancies between the new data and the existent network, it cannot suggest possible improvements; i.e., it cannot perform refinement.

## 2 The Refinement Problem

The refinement problem we address in this work is as follows: Given a set of new, partially specified, data, and an existent network structure, the objective is to produce a new, refined, network. The refined network should more accurately reflect the probabilistic structure of the new data, and at the same time retain as much of the old network structure as is consistent with the new data. The refinement process can be naturally viewed as a learning task. Specifically, the source data for the learning task consists of two pieces of information, namely the new data and the existent network structure. The goal of the learning task is to discover a more useful structure based on these two sources of information.

There are a number of features we desire of the learned network structure. First, the learned structure should accurately represent the distribution of the new data. Second, it should be similar to the existent structure. Finally, it should be as simple as possible. The justification of the first feature is obvious. The second arises due to the fact that the task is refinement, which carries with it an implicit assumption that the existent network is already a fairly useful, i.e., accurate, model. And the last feature arises from the nature of Bayesian network models: simpler networks are conceptually and computationally easier to deal with (see [LB92] for a more detailed discussion of this point).

It is easily observed that in many circumstances these requirements cannot be fulfilled simultaneously. For example, a learned structure that can accurately represent the distribution of the new data may possess a large topological deviation from the existent network structure, or it may have a very complex topological structure. On the other hand, a learned structure that is close to the existent structure might represent a probability distribution that is quite inaccurate with respect to the new data. Thus, there are tradeoffs between these criteria. In other words, the learned network structure should strike a balance between its accuracy with respect to the new data, its closeness to the existent structure, and its complexity. The advantage of employing the MDL principle in this context is that it provides an intuitive mechanism for specifying the tradeoffs we desire.

### 2.1 The form of the new data

We are concerned with refining a network containing $n$ nodes. These nodes represent a collection of domain variables $\vec{X} = \{X_1, \ldots, X_n\}$, and the structure and parameters of the network represents a distribution over the values of these variables. Our aim is to construct a refinement algorithm that can refine parts of the original network using new data that is partial. Specifically, we assume that the data set is specified as a $p$-cardinality table of cases or records involving a subset $\vec{X}_p$ of the variables in $\vec{X}$ (i.e., $\vec{X}_p \subseteq \vec{X}$ and $||\vec{X}_p|| = p \leq n$). Each entry in the table contains an instantiation of the variables in $\vec{X}_p$, the results of a single case. For example, in a domain where $n = 100$, suppose that each variable $X_i$ can take on one of the five values $\{a, b, c, d, e\}$. The following is an example of a data set involving only 5 out of the 100 variables.

| $X_4$ | $X_{12}$ | $X_{20}$ | $X_{21}$ | $X_{45}$ |
|---|---|---|---|---|
| b | a | b | e | c |
| a | d | b | c | a |
| b | b | a | b | b |
| ⋮ | ⋮ | ⋮ | ⋮ | ⋮ |
| ⋮ | ⋮ | ⋮ | ⋮ | ⋮ |

Using this data set we can hope to improve the original network by possibly changing the structure (the arcs) between the variables $X_4$, $X_{12}$, $X_{20}$, $X_{21}$, and $X_{45}$. The data set could possibly be used to improve the rest of the network also, if we employed techniques from missing data analysis, e.g., [SDLC93]. However, here we restrict ourselves to refinments of the structure between the variables actually mentioned in the data set.

## 3 Our Approach

As mentioned above, we employ the MDL principle in our refinement algorithm. The idea is to first learn a *partial network structure* from the new data and the existent network via an MDL learning method. This partial network structure is a Bayesian network structure over the subset of variables contained in the new data. Thus, it captures the probabilistic dependencies or independencies among the nodes involved. Based on this partial network, we analyze and identify particular spots in the original network that can be refined.

The process of constructing the partial network structure is like an ordinary learning task. The new data contains information about only a partial set of the original network variables. However, it is complete with respect to the partial network structure; i.e., it contains information about every variable in the partial structure. Hence, constructing the partial structure is identical to a learning task of constructing a Bayesian network given a collection of complete data points and additionally an existent network structure over a superset of variables.

In our previous work, we developed a MDL approach for learning Bayesian network structures from a collection of complete data points [LB93]. Unlike many other works in this area, our approach is able to make a tradeoff between the complexity and the accuracy of the learned structure. In essence, it prefers to construct a slightly less accurate network if more accurate ones require significantly greater topological complexity. Our approach also employed a best-first search algorithm that did not require an input node ordering.

In this paper, we extend our learning approach to take into account the existent network structure. Through the MDL principle, a natural aggregation of the new data and the existent network structure is made during the learning process. At the same time, a natural compromise will be sought if the new data and the existent structure conflict with each other.

The MDL principle provides a metric for evaluating candidate network structures. A key feature of our approach is the localization of the evaluation of the MDL metric. We develop a scheme that makes it possible to evaluate the MDL measure of a candidate network by examing the local structure of each node.

### 3.1 Applying the MDL Principle

The MDL principle states that the best model to be learned from a set of source data is the one the minimizes the sum of two description (encoding) lengths: (1) the description length of the model, and (2) the description length of the source data *given the model*. This sum is known as the *total description (encoding) length*. For the refinement problem, the source data consists of two components, the new data and the existent network structure. Our objective is to learn a partial network structure $H_p$ from these two pieces of information. Therefore, to apply the MDL principle to the refinement problem, we must find a network $H_p$ (the model in this context) that minimizes the sum of the following three items:

1. its own description length, i.e., the description length of $H_p$,
2. the description length of the new data given the network $H_p$, and
3. the description length of the existent network structure given the network $H_p$.

The sum of the last two items correspond to the description length of the source data given the model (item 2 in the MDL principle). We are assuming that these two items are independent of each other given $H_p$, and thus that they can be evaluated separately. The desired network structure is the one with the minimum total description (encoding) length. Furthermore, even if we cannot find a minimal network, structures with lower total description length are to be preferred. Such structures are superior to structures with larger total description lengths, in the precise sense that they are either more accurate representations of the distribution of the new data, or are topologically less complex, or are closer in structure to the original network. Hence, the total description length provides a metric by which alternate candidate structures can be compared.

We have developed encoding schemes for representing a given network structure (item 1), as well as for representing a collection of data points given the network (item 2) (see [LB92] for a detailed discussion of these encoding schemes). The encoding scheme for the network structure has the property that the simpler is the topological complexity of the network, the shorter will be its encoding. Similarly, the encoding scheme for the data has the property that the closer the distribution represented by the network is to the underlying distribution of the data, the shorter will be its encoding (i.e., networks that more accurately represent the data yield shorter encodings of the data). Moreover, we have developed a method of evaluating the sum of these two description lengths that localizes the computation. In particular, each node has a local measure known as its *node description length.* In this paper, we use $DL_i^{old}$ to denote the measure of the $i$-th node.[1] The measure $DL_i^{old}$ represents the sum of items 1 and 2 (localized to a particular node $i$), but it does not take into account the existent network; i.e., it must be extended to deal with item 3. We turn now to a specification of this extension.

# 4 The Existent Network Structure

Let the set of all the nodes (variables) in a domain be $\vec{X} = \{X_1, \ldots, X_n\}$, and the set of nodes in the new data be $\vec{X}_p \subseteq \vec{X}$ containing $p \leq n$ nodes. Suppose the existent network structure is $H_n$; $H_n$ contains of all the nodes in $\vec{X}$. Through some search algorithm, a partial network structure $H_p$ containing the nodes $\vec{X}_p$ is proposed. We seek a mechanism for evaluating item 3, above; i.e., we need to compute the description length of $H_n$ given $H_p$.

To describe $H_n$ given that we already have a description of $H_p$, we need only describe the differences between $H_n$ and $H_p$. If $H_p$ is similar to $H_n$, a description of the differences will be shorter than a complete description of $H_n$, and will still enable the construction of $H_n$ given our information about $H_p$. To compute the description length of the differences we need only develop an encoding scheme for representing these differences.

What information about the differences is sufficient to recover $H_n$ from $H_p$? Suppose we are given the structure of $H_p$, and we know the following information:

- a listing of the reversed arcs (i.e., those arcs in $H_p$ that are also in $H_n$ but with opposite direction),
- the additional arcs of $H_n$ (i.e., those arcs in $H_n$ that are not present in $H_p$), and
- the missing arcs of $H_n$ (i.e., those arcs that are in $H_p$ but are missing from $H_n$).

It is clear that the structure of $H_n$ can be recovered from the structure of $H_p$ and the above arc information. Hence, the description length for item 3, above, can be taken to be simply the length of an encoding of this collection of arc information.

A simple way to encode an arc is to describe its two associated nodes (i.e., the source and the destination node). To identify a node in the structure $H_n$, we need $\log n$ bits. Therefore, an arc can be described using $2\log n$ bits. Let $r$, $a$, and $m$ be respectively the number of reversed, additional and missing arcs in $H_n$ with respect to a proposed network $H_p$. The description length $H_n$ given $H_p$ is then given by:

$$(r + a + m)2\log n. \tag{1}$$

Note that this encoding allows us to recover $H_n$ from $H_p$.

This description length has some desirable features. In particular, the closer the learned structure $H_p$ is to the existent structure $H_n$, in terms of arc orientation and number of arcs in common, the lower will be the description length of $H_n$ given $H_p$. Therefore, by considering this description length in our MDL metric, we take into account the existent network structure, giving preference to learning structures that are similar to the original structure.

Next, we analyze how to localize the description length of Equation 1. Each arc can be uniquely assigned to its destination node. For a node $X_i$ in $H_n$ let $r_i$, $a_i$ and $m_i$ be the number of reversed, additional, and missing arcs assigned to it given $H_p$. It can easily be shown that Equation 1 can then be localized as follows:

$$\sum_{X_i \in \vec{X}} (r_i + a_i + m_i)2\log n. \tag{2}$$

Note that each of the numbers $r_i$, $a_i$, $m_i$, can be computed by examining only $X_i$ and its parents. Specifically, at each node $X_i$ we need only look at its incoming arcs (i.e., its parent nodes) in the structure $H_n$ and compare them with its incoming arcs in the structure $H_p$, the rest of $H_n$ and $H_p$ need not be examined.

Based on new data, $\vec{X}$ can be partitioned into two disjoint sets namely $\vec{X}_p$ and $\vec{X}_q$, where $\vec{X}_q$ is the set of nodes that are in $\vec{X}$ but not in $\vec{X}_p$. Equation 2 can hence be expressed as follows:

$$\sum_{X_i \in \vec{X}_p} (r_i + a_i + m_i)2\log n + \sum_{X_i \in \vec{X}_q} (r_i + a_i + m_i)2\log n.$$

The second sum in the above equation specifies the description lengths of the nodes in $\vec{X}_q$. Since these nodes are not present in the new data (i.e., they are not in $\vec{X}_p$), the corresponding $r_i$'s and $m_i$'s must be 0. Besides, the $a_i$'s in the second sum are not affected by the partial network structure $H_p$. That is, if we are searching for a good partial network, this part of the sum will not change as we consider alternate networks. As a result, the localization of the description length of the existent network structure (i.e., Equation 1) is

[1] This was denoted as simply $DL_i$ in our previous paper [LB93].

given by:

$$\mu + \sum_{X_i \in \vec{X}_p} (r_i + a_i + m_i) 2 \log n \qquad (3)$$

where $\mu$ is a constant that can be ignored when comparing the total description lengths of candidate network structures.

## 5 Learning the Partial Network Structure

Given the new data and an existent network structure, a partial network structure can be learned via the MDL principle by searching for a network with low total description length. The search algorithm evaluates the total description length of candidate networks using this to guide its search. As pointed out in Section 3.1, in previous work [LB93] we have been able to localize the computation of the first two components of the description length, generating a node measure function $DL^{old}$. Similarly, in the previous section we have shown how the third component of the description length can be localized. Combining these results, we introduce a new *node description length* measure for the refinement task. This is a local measure that assigns a weight to each node, and can be computed by examining only the node and its parents. The total description length of the network is then computed by simply summing over the nodes.

**Definition 5.1** The *node description length* $DL_i$ for the node $X_i$ is defined as:

$$DL_i = DL_i^{old} + (r_i + a_i + m_i) 2 \log n, \qquad (4)$$

where $DL_i^{old}$ is the local measure given in our previous work [LB93].

Note that any constant terms can be dropped as they do not play a role in discriminating between alternate proposed partial network structures $H_p$. Now the total description length of a proposed network structure $H_p$ is simply (modulo a constant factor) the $\sum_{X_i \in \vec{X}_p} DL_i$.

To obtain the desired partial network structure, we need to search for the structure with the lowest total description length. However, it is impossible to search every possible network: there are exponentially many of them. A heuristic search procedure was developed in our previous work [LB93] that has preformed successfully even in fairly large domains. This search algorithm can be applied in this problem to learn partial network structures by simply substituting our new description length function $DL$ for the old one $DL^{old}$

## 6 Refining Network Structures

Once we have learned a good partial network structure $H_p$ we can refine the original network $H_n$ by using information contained in $H_p$. The manner in which $H_p$ can be used is based on the following theorem about the node description length measure.

Let $\vec{X} = \{X_1, X_2, \ldots, X_n\}$ be the nodes in an existent Bayesian network $H_n$, $\vec{X}_p$ be any subset of $\vec{X}$, and $DL_{\vec{X}_p} = \sum_{X_i \in \vec{X}_p} DL_i$, where $DL_i$ is defined by Equation 4. Suppose we find a new topology for the subgraph formed by the nodes in $\vec{X}_p$ such that this new topology does not create any cycles when substituted into $H_n$; i.e., we find a new way to connect nodes in $\vec{X}_p$ that does not create any cycles with the rest of $H_n$. This new topology will alter the node description lengths of the nodes in $\vec{X}_p$. Let $DL^{new}_{\vec{X}_p}$ be the sum of the node description lengths of $\vec{X}_p$ under this new topology. Let $H_n^{new}$ denote the new network formed from $H_n$ by replacing the connections between the nodes in $\vec{X}_p$ by their new topology.

**Theorem 6.1** *If* $DL^{new}_{\vec{X}_p} < DL_{\vec{X}_p}$ *then* $H_n^{new}$ *will have a lower total description length than* $H_n$.

A proof of this theorem is given in [Lam94].

This theorem says that we can improve a network (i.e., find one with a lower description length) by improving one of its subgraphs. The only restriction is that the resulting network must remain acyclic. The theorem demonstrates the importance of our localization of the total description length metric into a node description length metric. The node description length metric allows us to refine a particular part of the network without having to evaluate the total description length of the entire network; a potentially expensive task if the network is very large.

Despite the fact that the new data only mentions a subset $\vec{X}_p$ of observed nodes from $\vec{X}$, it still represents a probability distribution over the nodes in $\vec{X}_p$. Hence, it contains information about the probabilistic dependencies or independencies among the nodes in $\vec{X}_p$, and as we have demonstrated, a partial network structure $H_p$ can be learned from the new data and the original network. In general, the structure $H_p$ is not a subgraph of the original network. Nevertheless, it contributes a considerable amount of new information regarding the interdependencies among the nodes in $\vec{X}_p$. In some cases, $H_p$ provides information that allows us to refine the original network, generating a better network with lower total description length. An algorithm performing this task is discussed below. In other cases, it can serve as an indicator for locating particular areas in the existent network that show dependency relationships contradicting the new data. These areas are possible areas of inaccuracy in the original network. This issue of using new data to monitor a network will be explored in future work.

### 6.1 A Refinement Algorithm

Suppose the existent network structure is $H_n$, and the learned partial structure is $H_p$. The objective of the refinement process is to obtain a refined structure of lower total description length (hopefully minimum) with the aid of the existent structure $H_n$ and the partial structure $H_p$.

Say we have a node $X_m$ in $H_p$. In $H_p$ this node has some set of parents $Par(X_m, H_p)$, and a its description length $DL_i$ Defn. 4 in $H_p$ can be computed. In the existent network $H_n$, however, $X_m$ will in general have a different set of parents $Par(X_m, H_n)$ and a different description length. If $Par(X_m, H_n) \not\subset \vec{X}_p$, then these two description lengths are incomparable. In this case $X_m$ has a parent in $H_n$ that does not appear in the new data; hence the new data cannot tell us anything about the effect of that parent on $X_m$'s description length. We identify all of the nodes $X_m$ whose parents in $H_n$ are also in $H_p$ and call these the set of *marked* nodes.

Suppose for a certain marked node $X_m$, we decide to substitute the parents of $X_m$ in $H_n$ with the parents of $X_m$ in $H_p$. After the substitution, a new structure $H_{n1}$ is obtained. Usually the total description length of $H_{n1}$ can be calculated simply by adding the total description length of the old structure $H_n$ to the difference between the local description lengths of $X_m$ in $H_n$ and $H_p$. The new total description length of $H_{n1}$ can be evaluated in this way if the substitution of the parents of $X_m$ in $H_n$ does not affect the local description lengths of any other node in $H_n$. In fact, the only situation where this condition fails is when the parents of $X_m$ in $H_p$ contain a reversed arc (as compared to $H_n$). Under this circumstance, we need to consider the node $X_r$ associated with this reversed arc. If $X_r$ is also a marked node, we need to re-evaluate its local description length since it will be affected by the substitution of $X_m$'s parents. Recursively, we must detect any other marked nodes that are, in turn, affected by the change in $X_r$'s description length. It can be easily observed that these affected nodes must all be connected. As a result, we can identify a *marked subgraph unit* that contains only marked nodes and which can be considered together as an unit when the replacement is performed.

Actually, we can obtain the same subgraph unit if we had started off at any node in the subgraph due to the symmetrical nature of the influence between the nodes in the subgraph. For instance, returning to the previous example, if we considered $X_r$ first, we would have detected that the local description length of $X_m$ would be affected by the substitution of $X_r$'s parents. The process would have continued and we would have obtained the same subgraph.

Figure 1 shows the algorithm for the identification of a marked subgraph unit with respect to $X_m$. Initially, $Q$ is a set containing the single node $X_m$ and it grows as

```
construct-subgraph (Q,X_m,M)
Let R be a set of reversed arcs from X_m's parent
set in H_p
For each X_r in R
        M = M - {X_r}
        If X_r is "marked" and X_r ∉ Q then
                Q = Q ∪ {X_r}
                construct-subgraph (Q,X_r,M)
```

Figure 1: The Algorithm for Identification of a Marked Subgraph Unit

```
partition-into-subgraph (M, S⃗)
while M ≠ ∅
        X_m is a node from M
        M = M - {X_m}
        Q = {X_m}
        construct-subgraph (Q,X_m,M)
        S⃗ = S⃗ ∪ {Q}
```

Figure 2: The Algorithm for Identification of All Subgraph Units

the algorithm progresses. $Q$ will contain the required marked subgraph when the algorithm terminates. Initially, $M$ is a set containing some nodes that could be transferred to $Q$. It shrinks as the algorithm progresses and contains the remaining marked nodes that are not included in $Q$.

Now, we can identify all marked subgraph units in $H_p$. Parent substitution is to be done for all the nodes in the subgraph if this subgraph is chosen for refinement. A useful property of the subgraph is that the change in description length of each subgraph is independent of all other subgraphs. Figure 2 shows the algorithm for identifying all marked subgraph units in $H_p$. Initially $M$ contains all of the marked nodes and $\vec{S} = \emptyset$. All subgraph units will be contained in $\vec{S}$ when the algorithm terminates. $Q$ is a local variable containing the nodes for the current subgraph unit.

The refinement problem now is reduced to choosing appropriate subgraphs for which we should perform parent substitution in order to achieve a refined structure of lowest total description length. Although each subgraph substitution yields an independent reduction in description length, these substitutions cannot be preformed independently as cycles may arise.

We use best-first search to find the set of subgraph units that yields the best reduction in description length without generating any cycles. To assist the search task, we construct a list $\vec{S} = \{S_1, S_2, \ldots, S_t\}$ by ranking all subgraphs in ascending order of the benefit gained if parent substitution was to be performed using that subgraph. The OPEN list contains search elements which consist of two components $\langle H, S\rangle$, where $H$ is a refined network structure and $S$ is the next subgraph unit to be substituted into $H$. The elements in the OPEN list are ordered by the sum of the description length of $H$ and the benefit contributed by the subgraph unit $S$. The initial OPEN list consists of the search elements $\langle H_i, S_{i+1}\rangle$ where $H_i$ is obtained by substituting $S_i$ into the existent structure $H_n$ for $i = 1$ to $t-1$.

1. Extract the first element from the OPEN list. Let it be $\langle H, S_i\rangle$. Put $H$ on to the CLOSED list.
2. Construct a new refined structure $H_{new}$ by incorporating $S_i$ into $H$.
3. Insert the element $\langle H, S_{i+1}\rangle$ into the OPEN list. If $H_{new}$ is acyclic, we also insert the element $\langle H_{new}, S_{i+1}\rangle$ into the OPEN list.
4. Terminate if our resource limits are exceeded.

# 7 Experimental Results

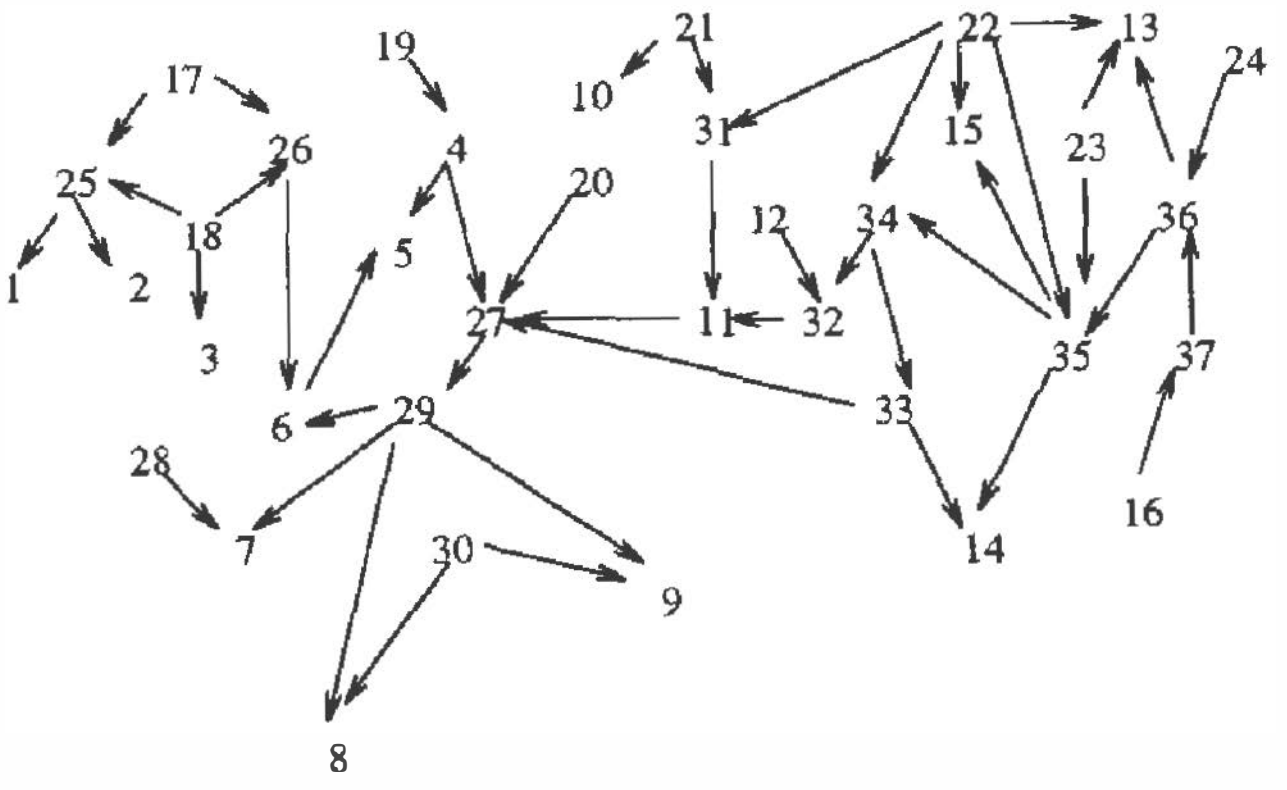

Figure 3: The True Structure Used To Generate The Data Set

Two experiments were conducted to verify our approach to refinement. The data used in these experiments was extracted from collection of complete data points that were generated from the ALARM structure shown in Figure 3. For the first two experiments, the new partially specified data was obtained by extracting from the complete data points the values of a subset of the variables. The extraction of partial data corresponds to performing a relational algebra "projection" on the full data set.

The first experiment used new data mentioning 17 nodes. The specific variables extracted for this experiment were variables number 1, 2, 3, 6, 7, 8, 9, 10, 17, 18, 21, 25, 26, 27, 28, 29, and 30.

The existent network structure before refinement was as shown in Figure 4. Note that we deliberately chose a slightly different topology from the correct one. The partial network structure recovered after the learning process is shown in Figure 5. Our refinement algorithm was then invoked. It succeeded in refining the existent network, Figure 4, so that it became identical to the true network, i.e., Figure 3, correcting all errors in the structure.

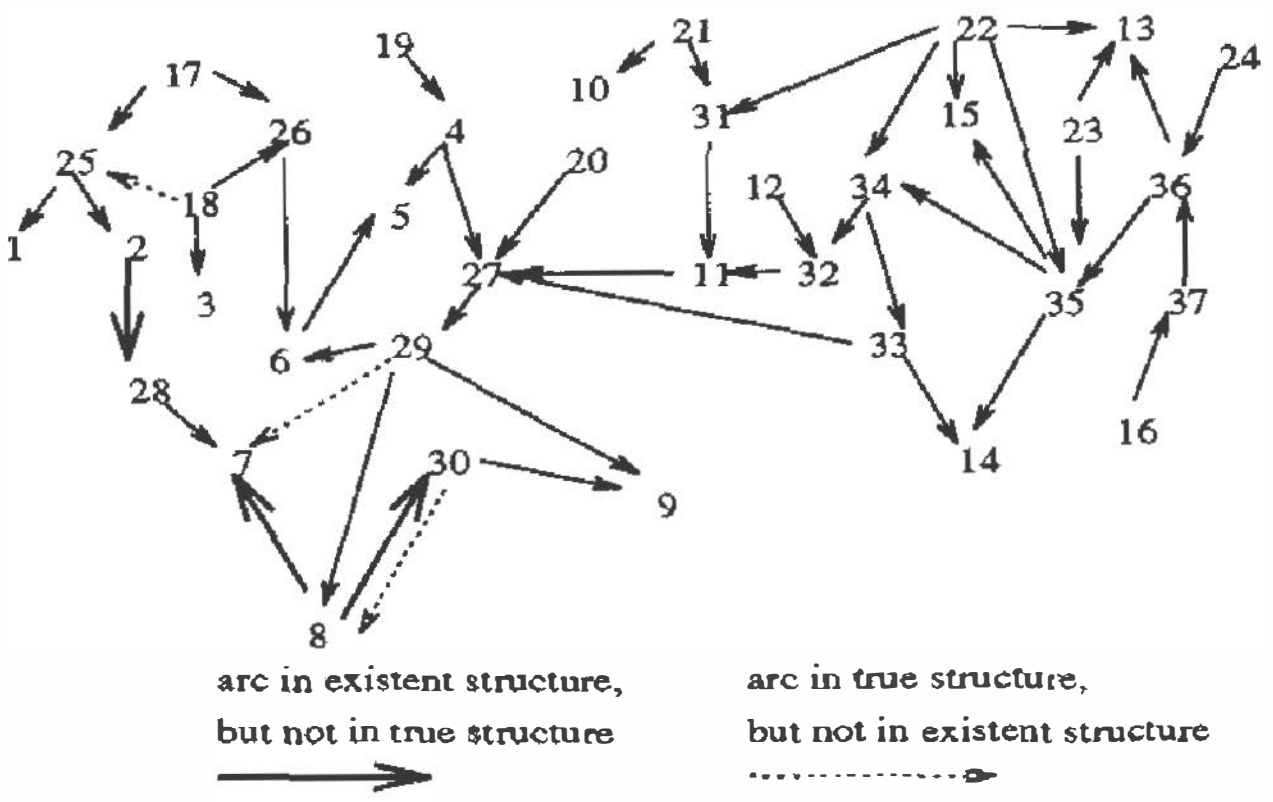

Figure 4: The Existent Structure for the First Experiment

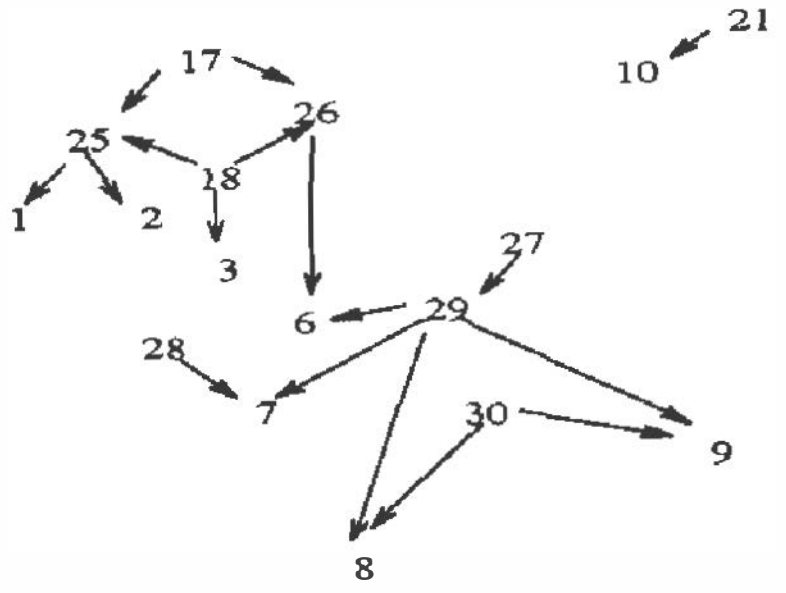

Figure 5: The Learned Partial Network Structure for the First Experiment

The second experiment used new data mentioning 24 nodes, specifically nodes 4, 5, 6, 10, 11, 13, 14, 16, 19, 20, 21, 22, 23, 24, 26, 27, 29, 31, 32, 33, 34, 35, 36, and 37. The existent network structure before refinement was shown in Figure 6. After the refinement process, the structure was improved to the point where it became identical to the true network, Figure 3, except for the arc between nodes 10 and 21, which remained reversed. This result, in fact, demonstrates the capability of our approach to optimize different features. If the issue of accuracy with respect to the new data was the only one considered, the arc connecting node 10 and 21 could be assigned in either direction: both directions yield the same accuracy. Any distribution that can be represented by the true structure, Figure 3, can be equally well represented by a structure in which the arc between 10 and 21 is reversed but is otherwise identical. This follows from the results of Verma and Pearl [VP90]. However, under the MDL metric used in our refinement approach, the direction from node 10 to 21 (i.e., 10 $\rightarrow$ 20) is preferred due to the bias from the existent structure, Figure 6. In other words, although accuracy with respect to the data is unable to decide a direction for this arc, the bias from the existent network makes our algorithm prefer to pre-serve the causal structure of the existent network if no information in the data contradicts this choice.

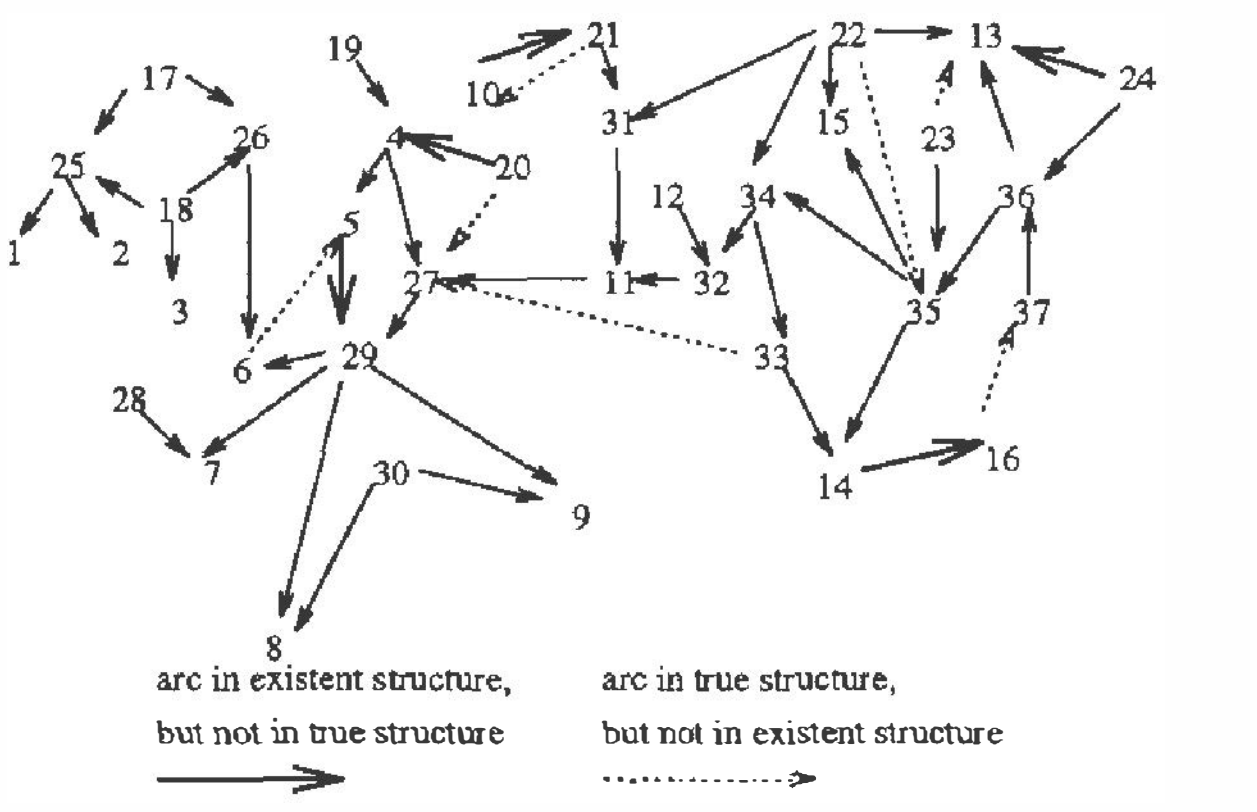

Figure 6: The Existent Structure for the Second Experiment

## Acknowledgments

We would like to thank E. Herskovits and M. Singh for providing the ALARM network database, and the referees for some insightful comments.